\documentclass[11pt,a4paper]{article}
\usepackage[whole]{bxcjkjatype}
\usepackage[unicode]{hyperref}

\usepackage[hyperref]{emnlp2018}
\usepackage{times}
\usepackage{latexsym}

\usepackage{breakurl}
\usepackage{amsmath}
\usepackage{amsfonts}
\usepackage{bm}
\usepackage[dvips]{graphicx}
\usepackage{here}
\usepackage{multirow}

\aclfinalcopy 

\newcommand{\argmax}{\mathop{\mathrm{argmax}}}

\title{Training Neural Machine Translation using Word Embedding-based Loss}

\author{Katsuki Chousa, Katsuhito Sudoh, Satoshi Nakamura \\
  Nara Institute of Science and Technology, 8916-5 Takayama-cho, Ikoma, Nara 630-0192, Japan \\
  {\tt \{k-chousa, sudoh, s-nakamura\}@is.naist.jp}\\
}
\date{}

\begin{document}
\maketitle
\begin{abstract}
In neural machine translation (NMT),
the computational cost at the output layer
increases with the size of the target-side vocabulary. 
Using a limited-size vocabulary instead may cause a significant decrease in translation quality.
This trade-off is derived from a softmax-based loss function that handles in-dictionary words independently,
in which word similarity is not considered.
In this paper, we propose a novel NMT loss function
that includes word similarity in forms of distances in a word embedding space.
The proposed loss function encourages an NMT decoder to generate words close to their references in the embedding space;
this helps the decoder to choose similar acceptable words
when the actual best candidates are not included in the vocabulary
due to its size limitation.
In experiments using ASPEC Japanese-to-English and IWSLT17 English-to-French data sets,
the proposed method showed improvements against a standard NMT baseline
in both data sets;
especially with IWSLT17 En-Fr, it achieved
up to +1.72 in BLEU and +1.99 in METEOR.
When the target-side vocabulary was very limited to 1,000 words,
the proposed method demonstrated a substantial gain, +1.72 in METEOR
with ASPEC Ja-En.
\end{abstract}

\section{Introduction}
\begin{table}[tb]
    \centering
    \begin{tabular}{rp{0.6\columnwidth}}\hline
        \textbf{source}: & 彼は古来まれな大政治家である。 \\ \hline
        \textbf{target}: & he is as great a statesman as ever lived . \\ \hline
        \textbf{attn-seq2seq}: & he is as great a \textless unk\textgreater\ as ever lived . \\ \hline
        \textbf{proposed}: & he is as great a politician as ever lived . \\ \hline
    \end{tabular}
    \caption{Example of translation from Japanese to English: an attn-seq2seq model generates \textless unk\textgreater\ token instead of \textit{statesman} because \textit{statesman} is an OOV word in this case. However, the proposed method can generate \textit{politician} instead of \textit{statesman}. It is easy to understand an outline of the sentence.}
    \label{table:example}
\end{table}

In neural machine translation (NMT) \citep{sutskever2014sequence, bahdanau2014neural, luong-pham-manning:2015:EMNLP},
the computational cost of the output layer,
in which generation probabilities of the output words are calculated,
increases with the size of the target language vocabulary.
Limiting the vocabulary size by ignoring low-frequency words
helps to reduce the cost,
but it also causes a significant decrease in the translation quality by the large number of out-of-vocabulary (OOV) words.
This trade-off is derived from a loss function to train NMT models, \textit{softmax cross-entropy}.
Softmax cross-entropy is a standard loss function for multi-class classification problems over multinomial distributions
to encourage giving a large probability mass to a correct class.

Suppose we are to optimize parameters to generate a reference word \textit{see} at a certain position.
Then, the probability of \textit{see} at that position will increase towards one,
and those of the other words will decrease towards zero regardless their meanings.
That is also the case even for some words similar to \textit{see},
such as \textit{look};
the occurrence of \textit{look} is penalized as well as other dissimilar words.
In this paper,
we argue that such penalty should be small for similar words and vice versa;
the occurrence of \textit{look} should be penalized much less than dissimilar words like \textit{listen}.
This problem becomes more serious when a reference word falls into
an OOV word due to its few occurrences in the training data.
In such cases with a typical implementation of NMT,
parameters are optimized to generate a special OOV symbol at that point
by penalizing the occurrences of all the other words.
Generating possible similar words there would be beneficial in practice,
compared with useless OOV symbols.
We focus on that problem in this work
and aim to implement a better loss function
according to the discussion above.

Word embeddings are continuous representations of words in a vector space.
Some different methods have been proposed such as word2vec \citep{mikolov2013efficient, mikolov2013distributed}, Glove \citep{pennington2014glove}, and FastText \citep{bojanowski2017enriching}.
An important finding by them is that
the vectors of syntactically and semantically similar words are close to each other.
That is very useful for introducing syntactic and semantic similarity
into an NMT loss function as discussed above.
The word embeddings can be obtained by using a large monolingual corpus,
also independently from a bilingual corpus for training NMT models.
\citet{qi18naacl} demonstrated an advantage of the use of word embeddings in pre-training,
but it basically works for better parameter initialization
and does not tackle our problem.

In this paper,
we propose a novel loss function for NMT
that gives a small penalty to similar words
and also a large penalty to dissimilar words.
The loss function is defined as a weighted average of distances
between word vectors of a reference word and the others in the target language vocabulary,
and those weights are given by the generation probabilities in its softmax layer.
Thus, the loss function explicitly penalizes undesirable word generation
of dissimilar words with high probabilities
and also encourages similar words to have large probabilities.
That nature is beneficial in situations
where many words fall into OOV in the translation results.
NMT models optimized by the proposed loss function
try to generate similar in-vocabulary words,
not just backing-off to special OOV symbols.
Table \ref{table:example} shows such an example.

We conducted experiments with two corpora: ASPEC and IWSLT17.
Experimental results show our models achieve consistent improvement on two language pairs over a standard NMT baseline.
In addition, our models with the small-size target-side vocabulary, which increase OOV words, also achieve significant improvement measured by a METEOR score of +1.72 points.

\section{NMT by Encoder-Decoder Model with Attention}
First, we review a baseline NMT model called \textit{encoder-decoder with attention}
following the formulation by \citet{luong-pham-manning:2015:EMNLP}.

Given a source sentence $X$ and a target sentence $Y$ as follows:
\begin{align*}
    X &= \left\{ \bm{x}_1, \bm{x}_2, ..., \bm{x}_J \right\}, \\
    Y &= \left\{ \bm{y}_1, \bm{y}_2, ..., \bm{y}_I \right\},
\end{align*}
where $\bm{x}_j \in \mathbb{R}^{S \times 1}$ is a one-hot vector of $j$-th input word,
$J$ is the length of input sentence $X$,
$\bm{y}_i \in \mathbb{R}^{T \times 1}$ is a one-hot vector of $i$-th output word,
and $I$ is the length of output sentence $Y$.

The problem of translation from the source to the target language can be solved
by finding the best target language sentence $\hat{Y}$ that maximizes the conditional probability,
\begin{equation}
    \hat{Y} = \argmax_{Y} p(Y|X).
\end{equation}
The conditional probability is decomposed by the product of conditional word probabilities
given the source language sentence and preceding target language words:
\begin{equation}
    p_{\theta}(Y|X) = \prod_{j=1}^{I}{p_{\theta}(\bm{y}_j|\bm{y}_{<j},X)},
    \label{eq:decoder}
\end{equation}
where $\bm{y}_{<j}$ represents the target language words up to position $j$,
and $\theta$ indicates the model parameters.

The model is composed of an encoder and decoder that are both implemented using recurrent neural networks (RNNs);
the encoder converts source language words into a sequence of vectors,
and the decoder generates target language words one-by-one based on the conditional probability shown in the equation (\ref{eq:decoder}).
The details are described below.

\subsection{Encoder}
The encoder takes a sequence of source language words $X$ as inputs
and returns forward hidden vectors $\overrightarrow{h_j}$ $(1 \leq j \leq J)$ of the forward RNN:
\begin{equation}
    \label{eq:encoder_hidden}
    \overrightarrow{\bm{h}_j} = f(\overrightarrow{\bm{h}_{j-1}}, \bm{x}_j).
\end{equation}
Similarly, we can obtain backward hidden vectors $\overleftarrow{h_j}$ $(1 \leq j \leq J)$ of the backward RNN, in the reverse order.
These forward and backward vectors are concatenated to form source vectors $\bm{h}_j$ $(1 \leq j \leq J)$ as follows:
\begin{equation}
    \bm{h}_j = \left[ \overrightarrow{\bm{h}_j}; \overleftarrow{\bm{h}_j} \right].
\end{equation}

\subsection{Decoder}
The decoder takes source vectors as inputs and returns target language words one-by-one.
The decoder RNN starts with the initial hidden vector $\bm{h}_J$
(concatenated source vector at the end),
and generates target words in a recurrent manner using its hidden state and an output context.
The conditional output probability of a target language word $\bm{y}_i$ defined as follows:
\begin{equation}
    p_{\theta}(\bm{y}_i | \bm{y}_{<i}, X) = \text{softmax}\left(\bm{W}_s\bm{\tilde{d}}_i\right).
    \label{eq:output_probability}
\end{equation}
Here, $\bm{W}_s$ is a parameter matrix in the output layer.
$\bm{\tilde{d}}_i$ is a vector calculated as follows:
\begin{align}
    \bm{\tilde{d}}_i &= \text{tanh}(\bm{W}_c\left[\bm{c}_i; \bm{d}_i\right]), \\
    \bm{d}_i &= g(\bm{d}_{i-1}, \bm{y}_{i-1}).
\end{align}
$\bm{W}_c$ is a parameter matrix, $g$ is an RNN function taking its previous state vector with the previous output word as inputs to update its state vector.
$\bm{c}_i$ is a context vector to retrieve source language inputs in forms of a weighted sum of the source vectors $\bm{h}_j$, defined as follows:
\begin{equation}
    \bm{c}_i = \sum_{j=1}^{S}{\alpha_{ij}\bm{h}_j}.
\end{equation}
$\alpha_{ij}$ is a weight for $j$-th source vector at the time step $i$ to generate $\bm{y}_{i}$,
defined as follows.
\begin{equation}
    \alpha_{ij} = \frac{\exp\left(score\left(\bm{d}_i, \bm{h}_j\right)\right)}{\sum_{j'=1}^{J}{\exp{\left(score\left(\bm{d}_i, \bm{h}_{j'}\right)\right)}}}.
\end{equation}
The $score$ function above can be defined in some different ways as discussed by \citet{luong-pham-manning:2015:EMNLP}.
In this paper, we use \textit{dot} attention for this \textit{score} function calculated as follows:
\begin{equation}
    score\left(\bm{d}_i, \bm{h}_j\right) = \bm{d}_i^{\mathrm{T}}\bm{h}_j.
\end{equation}
This scalar product score basically means the decoder puts more weights (\textit{attention}) to source vectors close to its state vector $\bm{d}_i$.

\section{Loss Functions}
In this section, we first review a standard loss function, softmax cross-entropy, 
then propose a novel word embedding-based loss.

\subsection{Softmax Cross-Entropy}
Softmax cross-entropy loss is a commonly used loss function for multi-class classification including word generation in NMT, defined as follows:
\begin{equation}
    \label{eq:cross_entropy}
    \ell_{ent} = - \sum_{i=1}^{I}{\sum_{j=1}^{J}{\bm{y}_{ij}\log{p_{\theta}\left(\bm{y}_{ij}\mid\bm{y}_{<i}, X\right)}}},
\end{equation}
where $\bm{y}_{ij}$ is j-th element of the one-hot vector corresponding to i-th words of the target sentence.

\subsection{Proposed Loss Function: Word Embedding Loss}
As we discussed earlier, this standard loss function equally penalizes all words other than the reference word, even for similar words.
To avoid this problem,
we propose a novel loss function called \textit{word embedding loss} that gives a small loss to a similar word and vice versa.
The word embedding loss is defined as a weighted average of distances to a reference word in continuous vector space, and the weights are based on the word output probabilities in equation (\ref{eq:output_probability}).
\begin{align}
    \label{eq:dwe_loss}
    \begin{split}
        \hspace{2mm}\ell_{emb} &= \\
        &\hspace{-9mm}\sum_{i=0}^{I}{\sum_{k=0}^{K}{p\left(\bm{y}_{i}|\bm{y}_{<i}, X\right)d\left(E\left(V_k\right), E\left(\bm{y}_i\right)\right)}},
    \end{split}
\end{align}
where $V_k$ is k-th words in the target-side vocabulary, $E(w)$ denotes a vector of word $w$ on word embeddings. A function {\it d} calculates the distance between two word vectors
and we use a Euclidean distance:
\begin{equation}
    d\left(\bm{s}, \bm{t}\right) = {\|\bm{s}-\bm{t}\|}.
\end{equation}

\section{Experiments}
We conducted three different experiments to examine the effects of the proposed method,
comparing various training strategies (\ref{subseq:strategy}), using a small-sized target language vocabulary (\ref{subsec:small}),
and different language pairs (\ref{subsec:language}).

\subsection{Settings}
All NMT models were implemented as described in the previous sections
using primitiv.\footnote{\url{https://github.com/primitiv/primitiv}}
Both the encoders and decoders were two-layered LSTMs \citep{hochreiter1997long:LSTM},
and the decoders used global dot attention to the source vectors.
We used a bi-directional encoder \citep{bahdanau2014neural} and input feeding \citep{luong-pham-manning:2015:EMNLP}.
Both the number of dimensions in word embedding and hidden vectors were 512,
and the minibatch size was 64.
We used the most frequent 20,000 words as source-side vocabulary in the training data for source language.
Adam \citep{Kingma2015} was used for optimization with the default parameter ($\alpha = 0.001, \beta_1 = 0.9, \beta_2 = 0.999, \epsilon = 10^{-8}$).
Gradient clipping was set to 5, and weight decay was set to $10^{-6}$.
The dropout probability was set to 0.3.
The learning rate was adjusted by a decay factor of $1/\sqrt{2}$ when a validation loss in a training epoch was larger than that in the previous epoch.
Then, we chose the best parameters with the smallest validation loss.
For evaluation metrics,
we used BLEU \citep{papineni-EtAl:2002:ACL} as a de-facto standard
and METEOR \citep{denkowski:lavie:meteor-wmt:2014} to focus on synonyms.

\subsection{Datasets}
We used two different corpora for the experiments:
Asian Scientific Paper Excerpt Corpus (ASPEC) \citep{NAKAZAWA16.621} for a mid-scale Japanese-to-English task in all three experiments,
and IWSLT17\footnote{\url{http://workshop2017.iwslt.org/}}
for a small-scale English-to-French task in the third experiment.
Table \ref{table:corpora} shows their detailed statistics.

The English and French sentences were tokenized by the Moses tokenizer.\footnote{\url{ https://github.com/moses-smt/mosesdecoder/blob/master/scripts/tokenizer/tokenizer.perl}}
The Japanese portion was tokenized by KyTea \citep{neubig11aclshort}.
We filtered out the sentences whose number of tokens was more than 60 from the training set.

We also used word embeddings for the target languages, English and French.
The English word embeddings were from pre-trained ones,\footnote{\url{https://code.google.com/archive/p/word2vec/}}
trained using Google News dataset.
The French word embeddings were trained using Wikipedia dumps\footnote{\url{https://dumps.wikimedia.org}} with gensim.\footnote{\url{https://radimrehurek.com/gensim/}}
We used word2vec \citep{mikolov2013efficient, mikolov2013distributed} as a training method to get word embeddings.

\begin{table}[tbp]
    \centering
    \begin{tabular}{c|c||c|c|c}
        \multirow{2}{*}{Corpus} & \multirow{2}{*}{Lang.} & \multicolumn{3}{c}{Number of Sentence} \\ \cline{3-5}
        & & Train & Valid. & Test \\ \hline\hline
        ASPEC & Ja-En & 964k & 1790 & 1812 \\
        IWSLT17 & En-Fr & 226k & 890 & 1210 \\
    \end{tabular}
    \caption{Number of sentences for each corpora.}
    \label{table:corpora}
\end{table}

\subsection{Effect of Training Strategy}\label{subseq:strategy}
In the first experiment, we examined the effect of different training strategies using the cross-entropy ($\ell_{ent}$) and proposed ($\ell_{emb}$) loss functions.
Our interest here was the best practice for using them in NMT training.
We compared three different combinations of loss functions:
\begin{itemize}
    \item $\ell_{ent}$ only (baseline)
    \item $\ell_{emb}$ only
    \item $\ell_{ent} + \ell_{emb}$
\end{itemize}
We also examined pre-training with the baseline loss followed by training with the proposed loss.
Here, we note that the only use of $\ell_{emb}$ without pre-training did not work; the training was trapped into a poor local minimum.
We set the size of the target-side vocabulary to 10,000.

\begin{table*}[tb]
    \centering
    \begin{tabular}{c|c|c||l|l}
        \hline
        target vocab. & loss & pre-train & BLEU & METEOR \\ \hline\hline
        20,000 & $\ell_{ent}$ & None & 24.91 & 30.71 \\ \hline\hline
        \multirow{4}{*}{10,000}& $\ell_{ent}$ & None & 23.78 & 29.39 \\
        & $\ell_{ent}+\ell_{emb}$ & None & \textbf{24.75} (+0.97) & 29.93 (+0.54) \\
        & $\ell_{ent}+\ell_{emb}$ & $\ell_{ent}$ & \textbf{24.60} (+0.82) & 29.52 (+0.13) \\
        & $\ell_{emb}$ & $\ell_{ent}$ & \textbf{24.85} (+1.07) & 29.81 (+0.41) \\ \hline\hline
        \multirow{4}{*}{1,000} & $\ell_{ent}$ & None & 14.21 & 18.43 \\
        & $\ell_{ent}+\ell_{emb}$ & None & 14.35 (+0.14) & 18.66 (+0.23) \\
        & $\ell_{ent}+\ell_{emb}$ & $\ell_{ent}$ & \textbf{14.72} (+0.51) & 18.88 (+0.45) \\
        & $\ell_{emb}$ & $\ell_{ent}$ & \textbf{14.74} (+0.53) & 20.15 (+1.72) \\ \hline
    \end{tabular}
    \caption{Experimental results score with ASPEC Japanese-to-English parallel corpus measured by BLEU and METEOR (values in parentheses are difference of BLEU or METEOR gains against baseline system). The bold BLEU scores indicate the difference from
the baseline system is statistically significant ($p<0.01$).}
    \label{table:aspec_result}
\end{table*}

The middle row of Table \ref{table:aspec_result} shows the results in BLEU and METEOR.
All of the methods using $\ell_{emb}$ resulted in higher BLEU and METEOR scores than the baseline just using $\ell_{ent}$,
and the BLEU score was comparable to the baseline model using 20,000 words as target language vocabulary.
There are no significant differences among them.
These results suggest that the proposed loss function is effective with a relatively small target language vocabulary.

\subsection{Effect of Target-side Vocabulary Size}\label{subsec:small}
In the second experiment,
we tested a very small target language vocabulary
to examine the robustness in a limited condition.
Here, we set the target language vocabulary size to just 1,000.

The bottom row of Table \ref{table:aspec_result} shows the results. 
All of the methods using $\ell_{emb}$ also resulted in better BLEU and METEOR scores.
The method that uses only $\ell_{emb}$ after baseline pre-training
showed remarkable improvements of +1.72 points in METEOR.

These results suggest that the proposed method works
with a limited vocabulary condition.

\subsection{Effect of Language Pairs}\label{subsec:language}
\begin{table*}[tbp]
    \centering
    \begin{tabular}{c|c|c||l|l}
        \hline
        target vocab. & loss & pre-train & BLEU & METEOR \\ \hline\hline
        \multirow{4}{*}{10,000} & $\ell_{ent}$ & None & 33.89 & 56.37 \\
        & $\ell_{ent}+\ell_{emb}$ & None & 33.94 (+0.05) & 57.20 (+0.83) \\
        & $\ell_{ent}+\ell_{emb}$ & $\ell_{ent}$ & \textbf{35.46} (+1.57) & 58.35 (+1.98) \\
        & $\ell_{emb}$ & $\ell_{ent}$ & \textbf{35.60} (+1.72) & 58.35 (+1.99) \\ \hline
    \end{tabular}
    \caption{Effect of language pairs on BLEU and METEOR scores with IWSLT17 English-to-French parallel corpus (values in parentheses are difference of BLEU or METEOR gains against $\ell_{ent}$). The bold BLEU scores indicate the difference from
baseline system is statistically significant ($p<0.01$).}
    \label{table:en-fr}
\end{table*}

Finally,
we compared the results in another language pair to examine
whether the advantage of the proposed loss function depends on a specific language.
We conducted the same experiments with the IWSLT17 English-to-French data
with the target language vocabulary size of 10,000.

Table \ref{table:en-fr} shows the results for English-to-French,
which can be compared with the Japanese-to-English results in the middle row of Table \ref{table:aspec_result}.
The results were basically similar but the improvement by the use of $\ell_{emb}$ with the pre-training was much larger in
these training strategies.
As in the experiment with the Japanese-English parallel corpus, all of the methods using $\ell_{emb}$ improve translation accuracies on BLEU and METEOR metrics, especially using only $\ell_{emb}$ after $\ell_{ent}$ pre-training.
The BLEU gains for English-to-French translation are bigger than those for Japanese-to-English.
This result suggests that the proposed method is beneficial for not just one language pair.

\subsection{Discussion}
Our experimental results show the advantage of using the proposed loss function in NMT,
especially in generating similar words with the help of relaxed constraints in the loss function.

Table \ref{table:translation_example} shows two translation examples in the English-to-Japanese experiment with ASPEC with a vocabulary size of 10,000.
These examples are generated by using the baseline model and proposed model, which use only $\ell_{emb}$ after $\ell_{ent}$ pre-training.

In Example (1), the target sentence includes the word \textit{eyeball} replaced with the special token \textless unk\textgreater\ as an OOV word.
The translation result by the baseline model contains an OOV word that would mean \textit{eyeball}.
On the other hand, the model trained using the proposed loss function generated \textit{eye} instead of \textless unk\textgreater.
This suggests the proposed method enables reasonable word choice for low-frequency words.

In Example (2), the target sentence also includes \textless unk\textgreater\, which was originally \textit{moldings}.
The proposed method gave a paraphrase \textit{the extrusion form} using limited in-vocabulary words for the phrase \textit{shape of moldings}. 

\begin{table*}[tbp]
    \centering
    \begin{tabular}{rp{0.8\textwidth}}
         Example (1) & \\ \hline
         \textbf{source} (Ja): & そのため，実際の眼球の断面画像から有限要素メッシュを作成することを試みた。 \\ \hline
         \textbf{target} (En): & Therefore, preparation of the finite element mesh from the cross-sectional images of actual \textless unk:eyeball\textgreater\ was attempted. \\ \hline
         \textbf{baseline}: & Therefore, it was tried that the finite element mesh was made from the cross section image of actual \textless unk\textgreater. \\ \hline
         \textbf{proposed}: & Therefore, it was tried that the finite element mesh was made from the cross section image of actual eye. \\ \hline \\
         Example (2) & \\ \hline
         \textbf{source} (Ja): & リザーバ内流動パターンと押出物形態。 \\ \hline
         \textbf{target} (En): & The flow pattern in the reservoir and the shape of \textless unk:moldings\textgreater . \\ \hline
         \textbf{baseline}: & Flow patterns in reservoir and \textless unk\textgreater\ forms. \\\hline
         \textbf{proposed}: & The flow pattern in the reservoir and the extrusion form. \\ \hline
    \end{tabular}
    \caption{Translation Examples in Japanese-to-English translation with ASPEC.}
    \label{table:translation_example}
\end{table*}

\section{Related Work}
An approach with a similar motivation was proposed recently by \citet{Maha:ACL:2018};
it uses word vectors for smoothing a loss function in neural network-based language modeling (also evaluated with NMT).
Their method aims to optimize the conditional log-probability of augmented output sentences sampled from the reward distribution.
The rewards are defined based on the cosine similarity in a semantic word embedding space.
They significantly improved the results on image captioning and machine translation with the token-level and sequence-level rewards.
However, they mentioned that the token-level rewards bring smaller improvement on machine translation tasks, unlike the image-captioning experiment.

Our method was developed independently of theirs
and has a difference in the loss definition;
we aim to minimize a weighted sum of distances from a reference word in a vector space,
considering all the other words instead of some sampled words.
Therefore, our token-level loss significantly improved the results by calculating much heavier than \citet{Maha:ACL:2018}.

The use of subwords \citep{sennrich-haddow-birch:2016:P16-12:BPE, kudo:2018:subword} is another approach to reducing OOV words.
\citet{sennrich-haddow-birch:2016:P16-12:BPE} showed that a subword-based system achieves better performance than a word-based system for translating rare words.
However, it does not tackle our problem directly;
when we use softmax cross-entropy, the generation probability of the correct word there will only increase towards one and those of the other words will decrease towards zero regardless of their meaning.

\section{Conclusion}
In this paper,
we proposed a new loss function for NMT
using weighted average of distance between a reference word and all the other target language words in semantic space.
The experimental results show advantages of the proposed method
in translation accuracy in BLEU and METEOR
and in robust word choice considering semantic similarity
in a limited-vocabulary condition.
Future work includes efficient loss calculation over target language words, the use of different types of word embeddings other than word2vec
\citep{mikolov2013efficient,mikolov2013distributed},
and further detailed evaluation of this kind of NMT approach typically by subjective evaluation.

\bibliography{emnlp2018}
\bibliographystyle{acl_natbib_nourl}

\end{document}